\pdfoutput=1
\documentclass[runningheads]{llncs}

\usepackage{graphicx}
\usepackage{hyperref}
\hypersetup{
    colorlinks=true,
    linkcolor=blue,      
    urlcolor=blue,       
    citecolor=blue     
}
\usepackage{booktabs}
\usepackage{amsmath,amssymb}
\newcommand{\method}{GeoPACT}

\renewcommand{\thefootnote}{\fnsymbol{footnote}}

\begin{document}

\title{Geometry-anchored PET-aware multimodal pseudo-CT synthesis for whole-body attenuation correction: the BIC-MAC Challenge}
\titlerunning{Geometry-Anchored Multimodal Pseudo-CT Synthesis}

\author{
Xuan Loc Nguyen\inst{1}\textsuperscript{*} \and
Hoang-Loc Cao\inst{1}\textsuperscript{*}\textsuperscript{$\dagger$} \and
Truong Thanh Hung Nguyen\inst{2}\textsuperscript{$\dagger$} \and\\
Phuc Ho\inst{1} \and 
Phuc Truong Loc Nguyen\inst{3} \and\\
Nguyen Truong Toan To\inst{3} \and
Hung Cao\inst{2}
}

\authorrunning{Nguyen, Cao et al.}

\institute{
University of Science, VNU-HCM, Ho Chi Minh City, Vietnam
\and
Analytics Everywhere Lab, University of New Brunswick, Canada
\and
Friedrich-Alexander-Universit{\"a}t Erlangen-N{\"u}rnberg, Germany
}

\begingroup
\renewcommand{\thefootnote}{
  \ifcase\value{footnote}
  \or *
  \or $\dagger$
  \fi
}

\footnotetext[1]{These authors contributed equally to this work.}
\footnotetext[2]{Corresponding authors. Email: \texttt{chloc22@clc.fitus.edu.vn}, \texttt{hung.ntt@unb.ca}.}
\endgroup

\maketitle

\begin{abstract}
The BIC-MAC challenge targets whole-body pseudo-CT synthesis from NAC-PET, Dixon MRI, and a 2D topogram for CT-less PET attenuation correction. We propose \textbf{\method}, a geometry-anchored multimodal framework that uses NAC-PET as the spatial reference and incorporates topogram and MRI features through gated residual fusion. Absolute coordinates and whole-body conditioning support anatomically consistent patch-based prediction. Training combines attenuation-map supervision with a differentiable PET-response surrogate to reduce errors relevant to downstream PET reconstruction. Full-resolution pseudo-CT volumes are generated using sliding-window inference without requiring CT or PET labels at test time.

\end{abstract}

\section{Introduction}

Accurate PET reconstruction requires attenuation correction at 511~keV. In PET/CT, attenuation maps are commonly derived from CT, whereas PET/MR requires them to be estimated without a corresponding CT acquisition \cite{carney2006,ladefoged2017}. The {\renewcommand{\thefootnote}{$\ddagger$}
The \textbf{BIC-MAC challenge}\footnote{\url{https://bic-mac-challenge.github.io/}}} addresses this problem at whole-body scale by synthesizing 3D pseudo-CT from NAC-PET, Dixon MRI, and a 2D topogram, with predictions evaluated through both attenuation-map accuracy and downstream PET reconstruction.

Learning-based attenuation correction has been investigated using MRI- and PET-derived inputs \cite{torrado2019,leynes2018,hwang2019,dong2019,chen2025}. BIC-MAC further requires combining modalities with different dimensionality and spatial reliability: NAC-PET follows the target grid but has limited anatomical detail; the topogram provides an aligned projection without depth information; and Dixon MRI offers complementary anatomical contrast but may be locally misaligned. Across segmentation, PET enhancement, and cross-modality registration, recent methods have therefore explored modality-specific feature extraction with adaptive fusion to better exploit complementary information from heterogeneous imaging sources \cite{chung2026modality,yar2026m2diff,chen2023dusfe}. This motivates fusion strategies that preserve reliable spatial information while selectively incorporating complementary modalities.

We propose \method, a geometry-anchored multimodal framework for whole-body pseudo-CT synthesis. NAC-PET defines the spatial reference, while topogram and Dixon MRI features are incorporated through gated residual fusion. Absolute positional information and whole-body conditioning support anatomically consistent patch-based prediction. Training further combines attenuation-space supervision with a differentiable PET-response surrogate to reduce errors relevant to downstream PET reconstruction.

\begin{figure}[t]
    \centering
    \includegraphics[width=\linewidth]{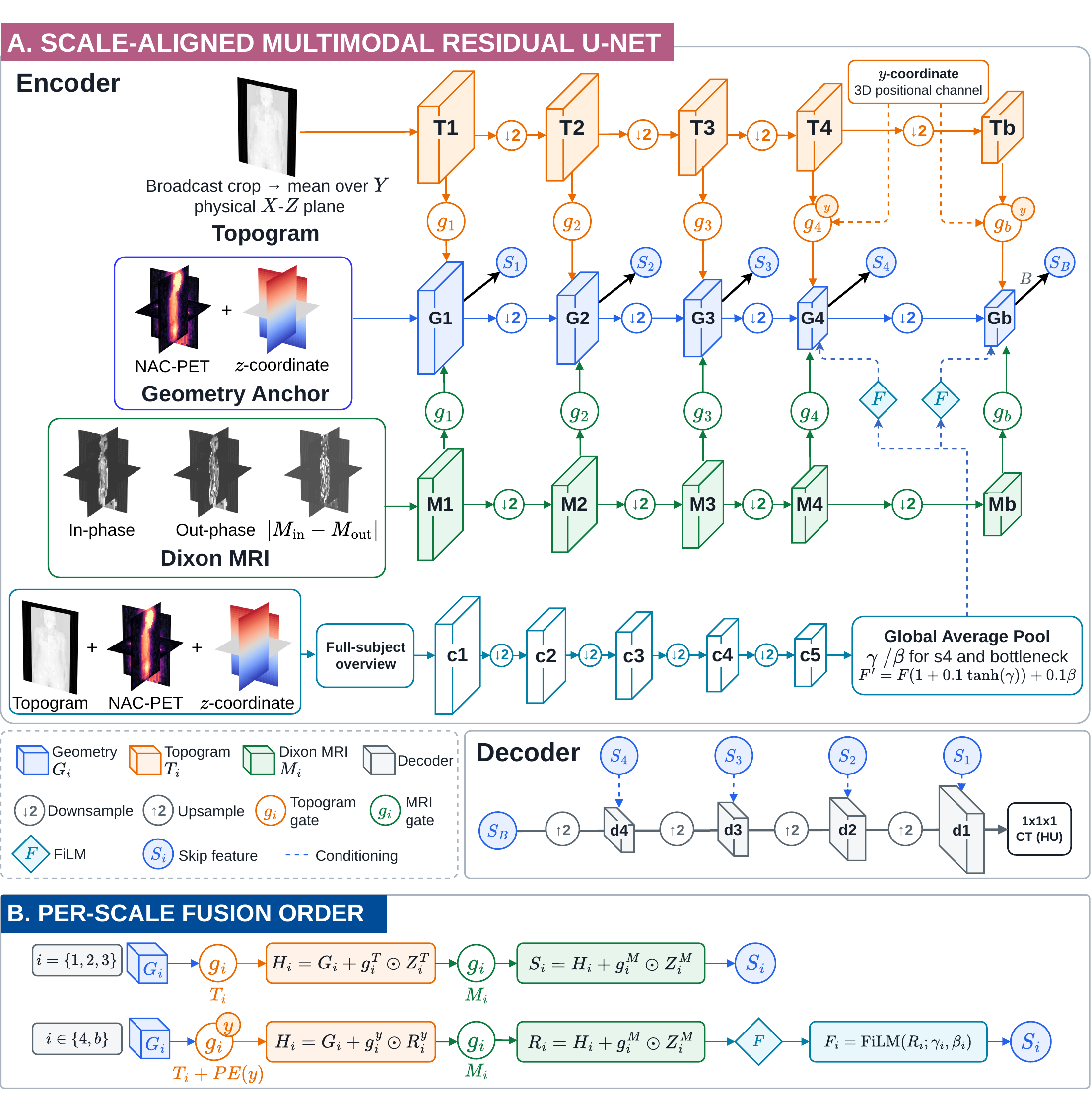}
    \caption{Overview of \method: geometry-anchored multimodal fusion with whole-body conditioning.}
    \label{fig:framework}
\end{figure}

\begin{figure}[h]
    \centering
    \includegraphics[width=\linewidth]{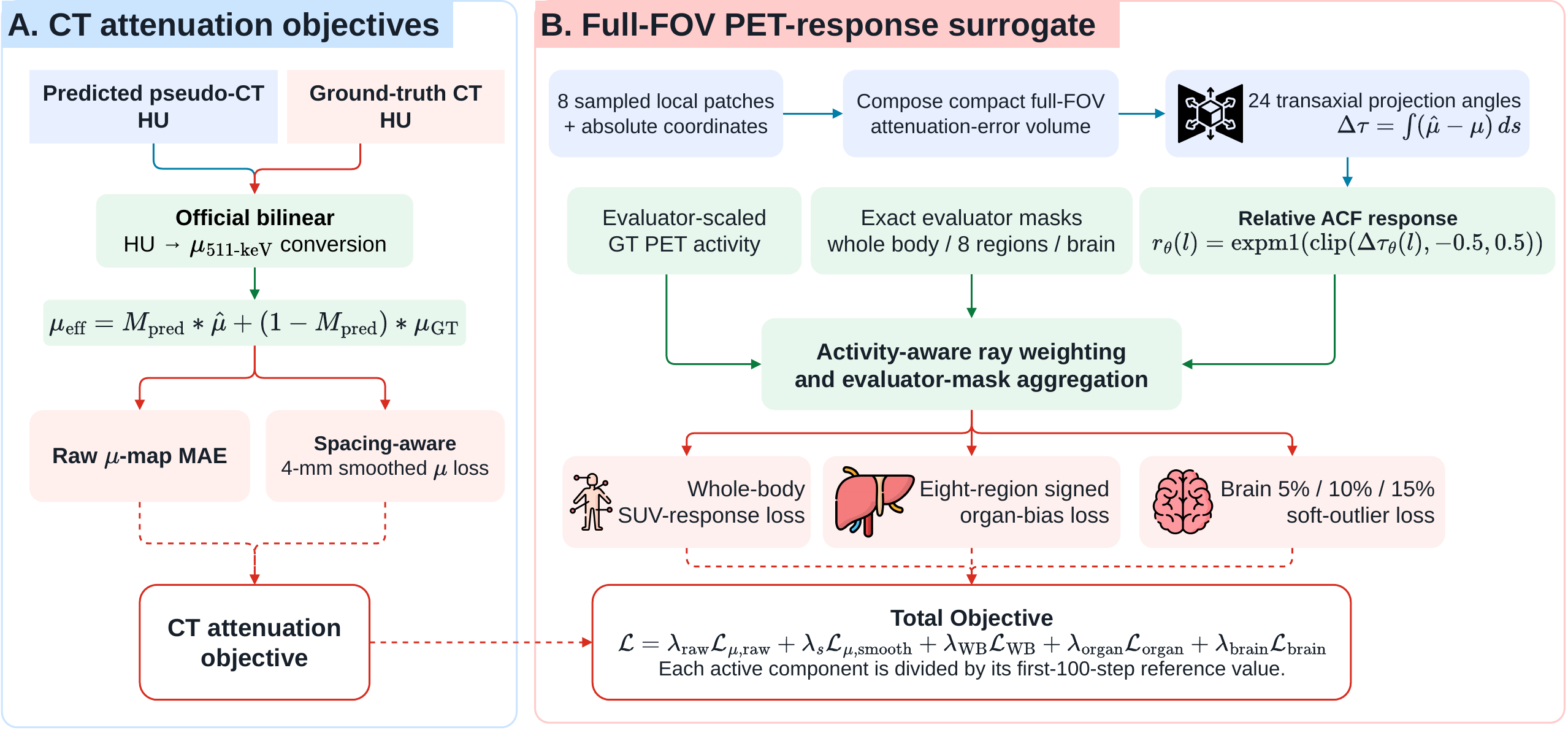}
    \caption{Training objectives of \method: (a) CT attenuation supervision and (b) the full-FOV PET-response surrogate}
    \label{fig:placeholder}
\end{figure}

\section{Method}
\label{sec:method}
\subsection{Preliminaries}
\label{sec:overview}
Let $P_{\mathrm{NAC}}$ denote the non-attenuation-corrected PET volume, $T$ the 2D topogram, and $M_{\mathrm{in}}$ and $M_{\mathrm{out}}$ the Dixon in-phase and out-of-phase MRI volumes.
Our model predicts a whole-body pseudo-CT $\hat C$ in Hounsfield units (HU),
\begin{equation}
\hat C
=
f_{\theta}\!\left(
P_{\mathrm{NAC}},
T,
M_{\mathrm{in}},
M_{\mathrm{out}},
c_y,
c_z
\right),
\label{eq:problem}
\end{equation}
where $c_y$ and $c_z$ denote the absolute anterior–posterior and cranio–caudal coordinates, respectively. The prediction is defined directly on the NAC-PET reference grid and preserves its NIfTI spatial metadata. Ground-truth CT, reconstructed CT-attenuation-corrected PET, and evaluator masks are used only during training.

The three input modalities provide complementary but differently reliable information. NAC-PET follows the target spatial grid but provides limited anatomical contrast. The topogram provides a stable whole-body projection but lacks depth resolution, whereas Dixon MRI provides complementary tissue contrast but may exhibit local spatial mismatch because of differences in acquisition geometry and patient pose. We therefore treat NAC-PET as the geometry anchor and allow topogram and MRI information to modify the geometry representation only through gated residual updates.

Figure~\ref{fig:framework} summarizes the complete architecture. A residual 3D U-Net processes NAC-PET together with the absolute cranio-caudal
coordinate $c_z$ and forms the geometry stream. A 2D topogram encoder and a lightweight 3D Dixon-MRI encoder generate scale-aligned auxiliary features. At each encoder level, topogram information is fused first, followed by MRI information. At the three finer scales $i\in\{1,2,3\}$, the resulting features $S_i$ are directly used as decoder skips. At the coarser scale $i=4$ and the bottleneck $i=b$, topogram fusion is additionally conditioned on the absolute anterior-posterior position, after which the fused features are modulated by subject-level whole-body context through feature-wise linear modulation (FiLM). The decoder starts from $S_b$, progressively combines the skip features $S_4,\ldots,S_1$, and predicts a single HU channel.

\subsection{Input Processing}
\label{sec:preprocessing}
All 3D modalities for a given subject are sampled using the same physical crop. Anatomical directions are determined from the NIfTI affine rather than from array indices. The topogram is represented in the physical $x$--$z$ plane and is broadcast along the physical $y$ direction only within its fusion branch.

Intensity normalization is computed once from the complete subject volume and then reused for all local patches. For modality $m$, values are first clipped to the $0.5$th and $99.5$th percentiles over valid voxels. NAC-PET is subsequently log-compressed, whereas MRI and topogram intensities use the identity transform. The normalized signal is:
\begin{align}
x_{\mathrm{norm}}^{(m)}
&=
\frac{
\phi_m\!\left(
\operatorname{clip}
\left(
x^{(m)},
q_{0.5}^{(m)},
q_{99.5}^{(m)}
\right)
\right)
-
\mu_s^{(m)}
}{
\sigma_s^{(m)}+\epsilon
},
\\
\phi_{\mathrm{NAC}}(x)
&=
\log\!\left(1+\max(x,0)\right),
\label{eq:normalization}
\end{align}
where $\mu_s^{(m)}$ and $\sigma_s^{(m)}$ are subject-level normalization statistics. This strategy avoids patch-dependent intensity scaling. The MRI branch receives three input channels: $M_{\mathrm{in}}$,
$M_{\mathrm{out}}$, and $|M_{\mathrm{in}}-M_{\mathrm{out}}|$.

Absolute coordinate channels are generated on the original subject grid as:
\begin{equation}
c(i)=\frac{2i}{N-1}-1,
\label{eq:coordinate}
\end{equation}
and are cropped together with each image patch. Consequently, a physical body location retains the same coordinate value during both random-patch training and sliding-window inference. The $z$-coordinate provides global body-region information to the geometry stream, whereas the $y$-coordinate disambiguates the depth of topogram features that have been lifted from a 2D projection into 3D space.

\subsection{Geometry-Anchored Multimodal Network}
\label{sec:network}
\paragraph{Geometry stream.}
The main pathway is a residual 3D U-Net with base width 24 and feature widths $24$, $48$, $96$, $192$, and $384$. It receives the normalized NAC-PET volume together with $c_z$. Each residual block contains two $3\times3\times3$ convolutions, instance normalization, leaky-ReLU activation, and either an identity or projected residual connection. Max pooling is used for downsampling, and dropout with a rate $0.2$ is applied at the bottleneck. The decoder uses transposed convolutions for upsampling, concatenates the corresponding fused encoder skip at each resolution, and applies another residual block. The complete model contains approximately $20.6$ million trainable parameters.

\paragraph{Topogram and MRI encoders.}
The topogram encoder is two-dimensional and follows the same five resolution levels as the geometry stream. At each level, its feature map is broadcast along the physical $y$ direction and projected to the corresponding geometry-channel width, yielding a scale-aligned topogram feature $T_i$.

At the three finer scales $i\in\{1,2,3\}$, $T_i$ is fused directly through a topogram gate $g_i^T$. At scale $i=4$ and the bottleneck $i=b$, the gate
additionally receives an eight-channel Fourier encoding of the absolute anterior--posterior coordinate, denoted $\operatorname{PE}(y)$. This
positional information helps the model determine how a depth-ambiguous 2D projection should influence each 3D location.

The MRI encoder receives $M_{\mathrm{in}}$, $M_{\mathrm{out}}$, and $|M_{\mathrm{in}}-M_{\mathrm{out}}|$. It forms an independent 3D branch with base width 12 and feature widths $12$, $24$, $48$, $96$, and $192$, producing scale-aligned features $M_i$. The reduced channel width limits memory consumption and prevents MRI information from directly redefining the NAC-PET reference geometry.

\paragraph{Sequential gated residual fusion.}
Let $G_i$, $T_i$, and $M_i$ denote the geometry, lifted-topogram, and MRI features at encoder level $i$, respectively. Following Module B in Figure~\ref{fig:framework}, fusion is sequential: the geometry feature is first updated using topogram evidence and is subsequently updated using MRI evidence.

At the three finer scales $i\in\{1,2,3\}$, the topogram gate is computed as:
\begin{equation}
g_i^T
=
\sigma\!\left(
\psi_i^T
\left(
[G_i,T_i,|G_i-T_i|]
\right)
\right),
\qquad i\in\{1,2,3\},
\label{eq:topo_gate_fine}
\end{equation}
and modulates the scale-aligned topogram residual $Z_i^T$:
\begin{equation}
H_i
=
G_i
+
g_i^T\odot Z_i^T,
\qquad i\in\{1,2,3\}.
\label{eq:fusion_fine_topo}
\end{equation}

MRI information is then incorporated through:
\begin{equation}
g_i^M
=
\sigma\!\left(
\psi_i^M
\left(
[H_i,M_i,|H_i-M_i|]
\right)
\right),
\qquad i\in\{1,2,3\},
\label{eq:mri_gate_fine}
\end{equation}
followed by:
\begin{equation}
S_i
=
H_i
+
g_i^M\odot Z_i^M,
\qquad i\in\{1,2,3\},
\label{eq:fusion_fine_mri}
\end{equation}
where $Z_i^M$ denotes the scale-aligned MRI residual correction. Thus, $H_i$ is the intermediate geometry representation after topogram fusion, and $S_i$ is the final fused feature passed to the decoder as the skip feature.

At the two coarsest levels $i\in\{4,b\}$, the topogram gate additionally uses the Fourier positional encoding $\operatorname{PE}(y)$:
\begin{equation}
g_i^{(y)}
=
\sigma\!\left(
\psi_i^T
\left(
[G_i,T_i,|G_i-T_i|,\operatorname{PE}(y)]
\right)
\right),
\qquad i\in\{4,b\}.
\label{eq:topo_gate_coarse}
\end{equation}
The corresponding geometry update is:
\begin{equation}
H_i
=
G_i
+
g_i^{(y)}\odot R_i^y,
\qquad i\in\{4,b\},
\label{eq:fusion_coarse_topo}
\end{equation}
where $R_i^y$ denotes the position-aware topogram residual correction shown in Figure~\ref{fig:framework}. MRI fusion is then applied using:
\begin{equation}
g_i^M
=
\sigma\!\left(
\psi_i^M
\left(
[H_i,M_i,|H_i-M_i|]
\right)
\right),
\qquad i\in\{4,b\},
\label{eq:mri_gate_coarse}
\end{equation}
and:
\begin{equation}
R_i
=
H_i
+
g_i^M\odot Z_i^M,
\qquad i\in\{4,b\}.
\label{eq:fusion_coarse_mri}
\end{equation}
Here, $R_i$ is the local multimodal feature after sequential topogram and MRI fusion and before whole-body conditioning.

The residual projections at the finer scales are zero-initialized, and the corresponding gates are biased toward small values. The coarse position-aware topogram residual is likewise initialized with a small amplitude. Consequently, the network initially remains close to the NAC-PET geometry pathway and learns to incorporate auxiliary evidence only when it is informative. This conservative design is particularly useful for Dixon MRI, where local misregistration with respect to NAC-PET may occur.

\paragraph{Whole-body context.}
Local patches cannot observe the full extent of the body. We therefore construct a compact full-subject overview from the topogram, NAC-PET, and absolute $z$-coordinate. The full-subject inputs are downsampled to $96\times96\times128$ and processed by a lightweight context encoder. Global average pooling produces a single subject-level context vector that predicts the FiLM parameters $(\gamma_i,\beta_i)$ for scale $i=4$ and the bottleneck $i=b$ \cite{film}.

Consistent with Figure~\ref{fig:framework}, FiLM is applied after both topogram and MRI fusion:
\begin{equation}
F_i
=
\operatorname{FiLM}
\left(
R_i;\gamma_i,\beta_i
\right)
=
R_i\odot
\left(
1+0.1\tanh\gamma_i
\right)
+
0.1\beta_i,
\qquad i\in\{4,b\}.
\label{eq:film}
\end{equation}

The conditioned features used by the decoder are therefore:
\begin{equation}
S_i
=
F_i,
\qquad i\in\{4,b\}.
\label{eq:conditioned_skip}
\end{equation}
The FiLM transformation is initialized to the identity, and the same whole-body context is reused for all local patches from a subject record.
Conditioning only the two coarsest levels provides global body-region information while preserving high-resolution local detail at the finer scales.

\subsection{Patch Training and Full-FOV Composition}
\label{sec:patches}
Training directly on full-resolution whole-body volumes is memory-intensive. Each subject record therefore contains eight patches of size $160\times160\times192$. Candidate patch centers are sampled from the prediction mask and randomly jittered along all three axes. A candidate is accepted when at least $35\%$ of the patch lies inside the prediction mask.

Patch sampling is stratified according to normalized physical body extent: lower body $0.00$--$0.22$, abdomen/pelvis $0.22$--$0.55$, thorax $0.55$--$0.78$, and head/neck $0.78$--$1.00$. Two patches are always sampled from the head/neck region, whereas the remaining six are sampled from the first three regions with relative weights $0.15$, $0.30$, and $0.30$, respectively.

Although the network is trained on local patches, the PET-aware objective requires attenuation information over long spatial ranges. For each subject record, the attenuation-error fields from the eight sampled patches are therefore placed into a compact $96\times96\times128$ full-field-of-view grid according to their absolute crop origins. Contributions from overlapping patches are averaged. Voxels not covered by any sampled patch are assigned zero attenuation error for that optimization step. This composition operation remains differentiable from the compact full-FOV response volume back to every covered local prediction.

\subsection{Attenuation and PET-Aware Objectives}
\label{sec:objectives}

Figure~\ref{fig:placeholder} summarizes the training objectives, which combine
direct attenuation supervision with a differentiable full-FOV PET-response
surrogate.

\paragraph{Attenuation-space losses.}
The official PET reconstruction operates on linear attenuation coefficients
rather than directly on CT Hounsfield units. We therefore convert both
predicted and reference CT to a 511-keV attenuation map using the official
bilinear transformation \cite{carney2006}. For $u=h+1000$,
\begin{equation}
\mu(h)
=
\begin{cases}
9.6\times10^{-5}u,
& u<1047,\\
5.10\times10^{-5}u+4.71\times10^{-2},
& u\geq1047.
\end{cases}
\label{eq:hu_to_mu}
\end{equation}
Predicted HU values are clipped to $[-1000,2000]$ before conversion, and
negative attenuation coefficients are subsequently clamped to zero.

We use two complementary attenuation-space losses. The first directly
measures masked attenuation error. The second follows the effective
attenuation input used for PET reconstruction: outside the prediction mask,
predicted attenuation is replaced by reference face-and-bed attenuation,
after which both prediction and target are smoothed using a spacing-aware
Gaussian kernel with $4$-mm full width at half maximum (FWHM).

Let $M_{\mathrm{CT}}$ denote the CT evaluation mask,
$M=M_{\mathrm{pred}}\cap M_{\mathrm{body}}$, and
$\mu_{\mathrm{water}}=0.096~\mathrm{cm}^{-1}$. The two losses are
\begin{align}
L_{\mu,\mathrm{raw}}
&=
\frac{
\sum_v
M_{\mathrm{CT},v}
\left|
\hat\mu_v-\mu_v
\right|
}{
\sum_v M_{\mathrm{CT},v}+\epsilon
},
\nonumber\\
L_{\mu,\mathrm{smooth}}
&=
\frac{1}{\mu_{\mathrm{water}}}
\frac{
\sum_v
M_v
\left|
\mathcal{G}_{4\mathrm{mm}}
(\hat\mu_{\mathrm{eff}})_v
-
\mathcal{G}_{4\mathrm{mm}}
(\mu)_v
\right|
}{
\sum_v M_v+\epsilon
},
\label{eq:attenuation_losses}
\end{align}
where
\begin{equation}
\hat\mu_{\mathrm{eff}}
=
M_{\mathrm{pred}}\hat\mu
+
(1-M_{\mathrm{pred}})\mu.
\label{eq:effective_mu}
\end{equation}
The raw loss preserves local attenuation accuracy, whereas the smoothed loss
emphasizes the effective attenuation field relevant to PET reconstruction.

\paragraph{Differentiable attenuation-response surrogate.}
PET attenuation correction depends on line integrals of the attenuation coefficient $\mu$ \cite{shi2019}. Instead of differentiating through a complete iterative PET reconstruction algorithm, we construct a differentiable surrogate that approximates the response to attenuation-map error.

Let $\delta\mu=\hat\mu-\mu$ denote attenuation error. For ray $l$ at transaxial projection angle $\theta$, we compute:
\begin{equation}
\Delta\tau_{\theta}(l)
=
\int_l
\delta\mu\,\mathrm{d}s,
\qquad
r_{\theta}(l)
=
\operatorname{expm1}\!\left(
\operatorname{clip}
\left(
\Delta\tau_{\theta}(l),
-0.5,
0.5
\right)
\right).
\label{eq:response}
\end{equation}

The compact full-FOV attenuation-error volume is projected at 24 transaxial angles ranging from $0^{\circ}$ to $172.5^{\circ}$ in increments of $7.5^{\circ}$. The resulting ray responses are backprojected and normalized by ray coverage to produce a voxelwise response field $R$. This field approximates the relative PET error induced by attenuation mismatch. For subjects with PET labels, ray weights are mildly adjusted using detached ground-truth PET activity and clipped to the range $[0.8,1.2]$. For subjects without PET labels, uniform ray weighting is used.

This module is an attenuation-response surrogate rather than a differentiable implementation of OSEM reconstruction. In particular, it does not explicitly model scatter, randoms, detector normalization, Poisson counting statistics, iterative activity updates, or the final reconstruction filter.

\paragraph{PET-aware terms.}
The response field $R$ defines three PET-aware losses that follow the
structure of the challenge evaluation. Let $A$ denote evaluator-scaled
activity, $M_{\mathrm{WB}}$ the whole-body evaluation mask, and $M_k$ the
mask corresponding to one of eight anatomical regions: brain, liver, spleen,
heart, pancreas, muscle, adipose tissue, and extremities.

The whole-body response loss is:
\begin{equation}
L_{\mathrm{WB}}
=
\frac{
\sum_v
M_{\mathrm{WB},v}
A_v
|R_v|
}{
\sum_v M_{\mathrm{WB},v}+\epsilon
}.
\label{eq:wb_loss}
\end{equation}
The regional term measures signed response bias within each anatomical region before taking its magnitude:
\begin{equation}
L_{\mathrm{organ}}
=
\frac{1}{8}
\sum_{k=1}^{8}
\left|
\frac{
\sum_v
M_{k,v}
A_v
R_v
}{
\sum_v
M_{k,v}
A_v+\epsilon
}
\right|.
\label{eq:organ_loss}
\end{equation}
Computing the signed regional mean before the absolute value preserves cancellation between positive and negative response errors within the same region.

The brain term $L_{\mathrm{brain}}$ follows the evaluator's brain-outlier criterion. Differentiable sigmoid approximations are used for the $5\%$, $10\%$, and $15\%$ relative-error thresholds, and the resulting soft outlier responses are averaged over the valid brain mask. Exact evaluator-scaled activity and evaluator masks are available for eight training subjects. For the remaining subjects, the detached NAC-PET activity, together with available CT-space masks, provides weaker fallback supervision. All 75 subjects contribute to the attenuation-space losses.

\subsection{Total Objective, Optimization, and Inference}
\label{sec:optimization}
The five loss terms operate at substantially different numerical scales. During the first 100 optimization steps for each active component, a reference magnitude is estimated as:
\begin{equation}
\operatorname{ref}_j
=
\frac{1}{100}
\sum_{n=1}^{100}
L_j^{(n)},
\qquad
\widetilde L_j
=
\frac{
L_j
}{
\operatorname{ref}_j+\epsilon
}.
\label{eq:loss_normalization}
\end{equation}
The complete objective is:
\begin{equation}
L_{\mathrm{total}}
=
\lambda_{\mu,r}
\widetilde L_{\mu,\mathrm{raw}}
+
\lambda_{\mu,s}
\widetilde L_{\mu,\mathrm{smooth}}
+
\lambda_{\mathrm{WB}}
\widetilde L_{\mathrm{WB}}
+
\lambda_{\mathrm{organ}}
\widetilde L_{\mathrm{organ}}
+
\lambda_{\mathrm{brain}}
\widetilde L_{\mathrm{brain}}.
\label{eq:total_loss}
\end{equation}

Training follows the curriculum in Table~\ref{tab:loss_schedule}. The initial stage establishes attenuation accuracy. PET-aware supervision is introduced only after the pseudo-CT prediction becomes stable, reducing the risk that the approximate response model dominates early optimization.

\begin{table}[t]
\centering
\caption{Loss-weight schedule. Each active component is divided by its
first-100-step reference value before weighting.}
\label{tab:loss_schedule}
\small
\setlength{\tabcolsep}{4pt}
\resizebox{.45\linewidth}{!}{
\begin{tabular}{@{}l|rrrrr@{}}
\toprule
\textbf{Epochs}
&
$\lambda_{\mu,r}$
&
$\lambda_{\mu,s}$
&
$\lambda_{\mathrm{WB}}$
&
$\lambda_{\mathrm{organ}}$
&
$\lambda_{\mathrm{brain}}$
\\
\midrule
$1$--$30$
& $1.00$
& $0.50$
& $0.00$
& $0.00$
& $0.00$
\\
$31$--$90$
& $1.00$
& $0.50$
& $0.05$
& $0.05$
& $0.02$
\\
$91$--$117$
& $1.00$
& $0.50$
& $0.15$
& $0.15$
& $0.05$
\\
$118$--$150$
& $1.00$
& $0.50$
& $0.20$
& $0.20$
& $0.015$
\\
\bottomrule
\end{tabular}
}
\end{table}

Each epoch contains 75 unique subject records together with one additional record for each of the eight PET-labelled subjects, resulting in 83 records per epoch. Repeated PET-labelled subjects resample their patch sets, thereby increasing the frequency of exact PET supervision without duplicating fixed
image crops.

The network is trained for 150 epochs using AdamW with an initial learning rate of $10^{-4}$, weight decay of $10^{-5}$, cosine learning-rate annealing, mixed-precision training, and gradient-norm clipping at 12. The batch size is one subject record, corresponding to eight local patches.

At inference, pseudo-CT is generated using a $160\times160\times192$ sliding window with $50\%$ overlap. Whole-body context is computed once for the complete subject and reused for every local window. Overlapping HU predictions are blended directly on the NAC-PET reference grid to obtain the final full-resolution pseudo-CT. No CT, reconstructed PET label, organ mask, or evaluator mask is required during inference.

The deployed model is obtained by interpolating the balanced V4.5 epoch-150 checkpoint with the PET-emphasized V4.5b epoch-35 checkpoint:
\begin{equation}
\theta_{\mathrm{final}}
=
(1-0.58)\,
\theta_{\mathrm{V4.5},e150}
+
0.58\,
\theta_{\mathrm{V4.5b},e35}.
\label{eq:model_interpolation}
\end{equation}
Only compatible floating-point model tensors are interpolated. The interpolation coefficient is selected using the official public reconstructed-PET validation results and is therefore treated as model selection rather than independent test evidence.

\section{Results}

All experiments were conducted on an NVIDIA DGX Spark equipped with the GB10 Grace Blackwell Superchip and 128 GB of unified system memory. Table~\ref{tab:public-results} summarizes the performance of \method~on the official four-case public validation set, compared with the
organizer-provided NAC-PET-only 3D U-Net baseline. Lower values indicate
better performance for all evaluation metrics.

\begin{table}[t]
\centering
\caption{Official public-validation results on the BIC-MAC challenge.
Lower is better for all metrics.}
\label{tab:public-results}
\small
\setlength{\tabcolsep}{3.5pt}
\renewcommand{\arraystretch}{1.20}

\resizebox{.9\linewidth}{!}{
\begin{tabular}{@{}lcccc@{}}
\toprule
\textbf{Method}
& \multicolumn{1}{c}{\textbf{CT}}
& \multicolumn{3}{c}{\textbf{PET}} \\
\cmidrule(lr){2-2}
\cmidrule(lr){3-5}

& \textbf{$\mu$-map MAE}
& \textbf{SUV MAE}
& \textbf{Organ Bias (\%)}
& \textbf{Brain Outlier} \\
\midrule

3D U-Net (baseline)
& 0.006610
& 0.0612
& 4.56
& 0.0541 \\

\textbf{\method\ (ours)}
& \textbf{0.005821}
& \textbf{0.0314}
& \textbf{2.23}
& \textbf{0.0053} \\

\bottomrule
\end{tabular}
}
\end{table}

\method~consistently outperforms the 3D U-Net baseline across all four evaluation metrics. Specifically, \method~achieves a CT $\mu$-map MAE of 0.005821 and an SUV MAE of 0.0314, corresponding to relative reductions of 11.9\% and
48.7\%, respectively. For the PET-based regional metrics, \method~reduces the organ bias from 4.56\% to 2.23\% and the brain outlier score from 0.0541 to 0.0053, yielding improvements of 51.1\% and 90.2\%, respectively. These results demonstrate that \method~provides consistent
improvements in both attenuation-map estimation and downstream PET quantification, with particularly substantial gains in the PET-based evaluation metrics.

\section{Conclusion}

We presented \method, a geometry-anchored approach for the BIC-MAC whole-body pseudo-CT synthesis task. The method treats NAC-PET as the spatial reference and incorporates topogram and Dixon MRI information through controlled residual fusion, complemented by positional and whole-body context. PET-aware training links attenuation-map prediction to an approximation of downstream reconstruction sensitivity while keeping inference simple and label-free. The framework is designed to address the central challenge of BIC-MAC by integrating modalities with different dimensionalities and spatial reliability for CT-less attenuation correction.

\section*{Acknowledgment}
This work is supported by NSERC Discovery Grant No RGPIN-2025-04478 and NSERC Discovery Supplement Award No DGECR-2025-00129.
The authors also thank the organizers of the BIC-MAC Challenge for providing the dataset, evaluation framework, and benchmarking platform used in this study.

\bibliographystyle{splncs04}
\bibliography{mybibliography}

\end{document}